\documentclass[11pt]{article}
\usepackage[a4paper, total={12cm, 20cm}]{geometry}
\usepackage[T1]{fontenc}
\usepackage{mathptmx}
\usepackage[utf8]{inputenc}
\usepackage[pdftex]{graphicx}
\usepackage{url}
\usepackage[font=small, labelfont=bf, justification=justified]{caption}
\usepackage[style=apa, backend=biber,natbib]{biblatex}
\providecommand{\keywords}[1]
{
  \small	
  \textbf{\textit{Keywords--}} #1
}

\title{Searching for Designs in-between}

\author{Camilo Cruz Gambardella \and Jon McCormack}

\date{}

\begin{document}
\maketitle
\begin{abstract}
The use of evolutionary methods in design and art is increasing in diversity and popularity. Approaches to using these methods for creative production typically focus either on optimisation or exploration. In this paper we introduce an evolutionary system for design that combines these two approaches, enabling users to explore landscapes of design alternatives using design-oriented measures of fitness, along with their own aesthetic preferences. We test our methods using a biologically-inspired generative system capable of producing 3D objects that can be exported directly as 3D printing toolpath instructions. For the search stage of our system we combine the use of the CMA-ES algorithm for optimisation and linear interpolation between generated objects for feature exploration. We investigate the system’s capabilities by evolving highly fit artefacts and then combining them with aesthetically interesting ones.
\end{abstract}


\section{Introduction}
\label{s:intro}

Evolutionary Computing (EC) methods have been used to address a wide variety of formal and functional problems in creative practice since the mid 1980s \citep[for an introduction][see]{frazer1995evolutionary}. The most common applications of these techniques are the optimisation of performance-oriented attributes of artefacts, and the exploration of novel design alternatives. For instance, in architecture, evolutionary methods, among other metaheuristics \citep{wortmann2016black}, are used in conjunction with parametric models \citep{woodbury2010elements} to search for efficient design alternatives, where efficiency is clearly defined through formal fitness measures. In contrast, visual artists and researchers in creative fields commonly use search methods that incorporate aesthetic selection mechanisms along with stochastic generative systems, to explore vast landscapes of alternatives looking for ‘interesting’ and novel phenotypes \citep{mccormack2005open}.

Typical design problems addressed with these techniques in architecture and engineering are the optimisation of structural elements \citep{goldberg1986engineering, jenkins1991towards}, the design of efficient building layouts \citep{jo1998space, wong2009evoarch}, sunlight and energy performance of buildings \citep{caldas2001evolution} and acoustic optimisation of concrete shells \citep{mendez2013multi}, to mention a few. All of these examples use EC methods to drive mostly deterministic parametric models, which generate alternative phenotypes. In other words, it is via the combination of these computational methods that the designed artefacts they produce acquire their characteristics.

Creative exploration using EC differs from optimisation-oriented approaches in that the main objective is not efficiency, but finding novel alternatives. In many cases artists substitute formal fitness functions with their own aesthetic judgement, putting a ``human in the loop'' of the evolutionary algorithm \citep{Dawkins1986}. Seminal examples in this field can be found in the work of William Latham and Karl Sims, who in the late 1980s began using evolutionary techniques, like selection and mutation, to generate complex forms and imagery \citep{sims1991artificial, Todd1992}, as well as to evolve Virtual Creatures \citep{sims1994evolving}. Another example is the work on aesthetic fitness and evolution by \citet{dorin2001aesthetic}, who explores the aesthetic capabilities of human-guided computational evolution via the generation of digital images. More recently, Dutch artists Erwin Driessens and Maria Verstappen’s Accretor \citep{whitelaw2015accretor} used evolutionary techniques to produce 3D shapes by evolving sets of rules for a cellular automaton. In this case the artists’ process involves monitoring the state of the evolving objects, in search of abrupt changes along their evolutionary trajectory. Then, the process is interrupted and the objects are fine-tuned manually. Another example is the work of artist Andy \citet{lomas2013}, who combines the use of a cellular division-inspired generative system with interactive genetic algorithms to generate aesthetically complex digital 3D objects, which he then categorises and evaluates aesthetically based on his own preference. Lomas’ latest work builds on his generate and evolve process, focusing on the exploration of the areas of the fitness landscape of his system located in-between categories by interpolating between objects, and looking for sudden/unexpected changes as the objects transition between states.

In this paper we introduce a prototype EC system for the creation of physical artefacts that combines evolutionary strategies (ES) with a novel, bio-inspired, agent-based generative system. By coupling a stochastic generative method capable of producing a wide range of diverse alternatives with a flexible ES driven by fitness functions that respond to holistic design objectives, our system integrates elements of creative exploration with design optimisation, enabling users to purposefully and holistically explore complex design spaces.

\begin{figure}[hbt]
    \centering
    \includegraphics[width=\textwidth]{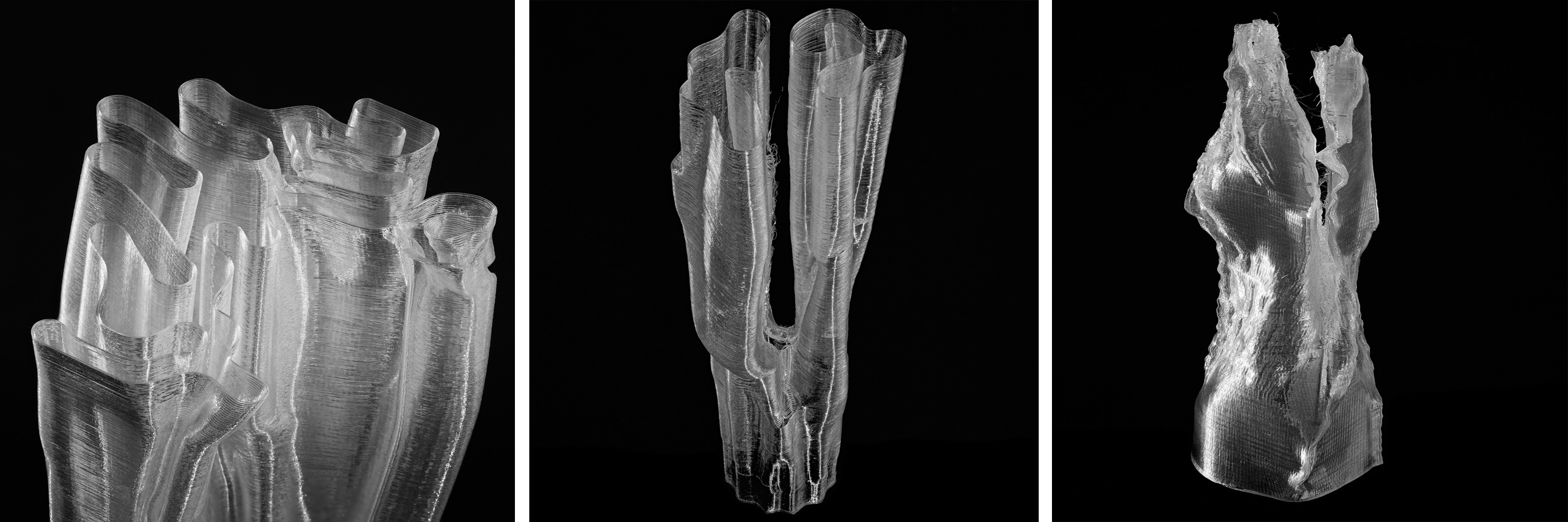}
    \caption{Examples of 3D printed forms grown using our generative system.}
    \label{fig:heroForm}
\end{figure}

\section{Generative System}
\label{s:generative-system}

The guiding principle behind the implementation of our generative system was the generation of interesting 3D forms that could be directly translated into digital fabrication instructions. For this purpose we defined a model that generates 2D shapes and transforms them over time. By capturing the state of these shapes at discrete time intervals as they develop, we obtain geometric information analogue to the representation used by slicing software to prepare 3D models for fused deposition modelling 3D printing (FDM): a series of stacked contours (Figure \ref{fig:heroForm}).

To drive the dynamics of the 2D shapes and produce complex geometry we borrowed concepts from natural differential growth processes \citep{barlow1989differential}, as seen in the work of Anders \citet{hoff2015}. We use the metaphor of ``organisms'': sequences of artificial cells connected to their neighbours to form closed loops. Cellular processes -- a rich source of complexity and diversity \citep{ball1999self} -- serve as inspiration for our system, directing the behaviours of our synthetic organisms. However, it is important to emphasise that the focus of our model is form-generation for creative purposes, and not the accurate reproduction of physical or biological phenomena.

\subsection{Building Blocks}
\label{ss:building-blocks}

Our generative system operates on the metaphor of \textit{colonies of organisms}, that develop over time in discrete timesteps. Each organism in a colony is defined as an array of cells $(C_1, C_2 \ldots C_n)$ connected by edges $(e_1, e_2 \ldots e_n)$, where $e_i = (C_i, C_{i+1})$ and $e_n = (C_n, C_1)$, forming closed loops.

Colonies develop in a synthetic environment -- a simulated viscous medium of uniform density. Sources of nutrients are randomly distributed across the environment. Each source has a limited amount of nutrition units, which are released, one at the time, every timestep, and diffused over their neighbouring area. Once nutrients have been released they progressively loose nutritional value over time. Once the source of nutrients is fully depleted, it disappears, giving rise to a new source randomly positioned within the development environment.

\begin{figure}[hbt]
    \centering
    \includegraphics[width=\textwidth]{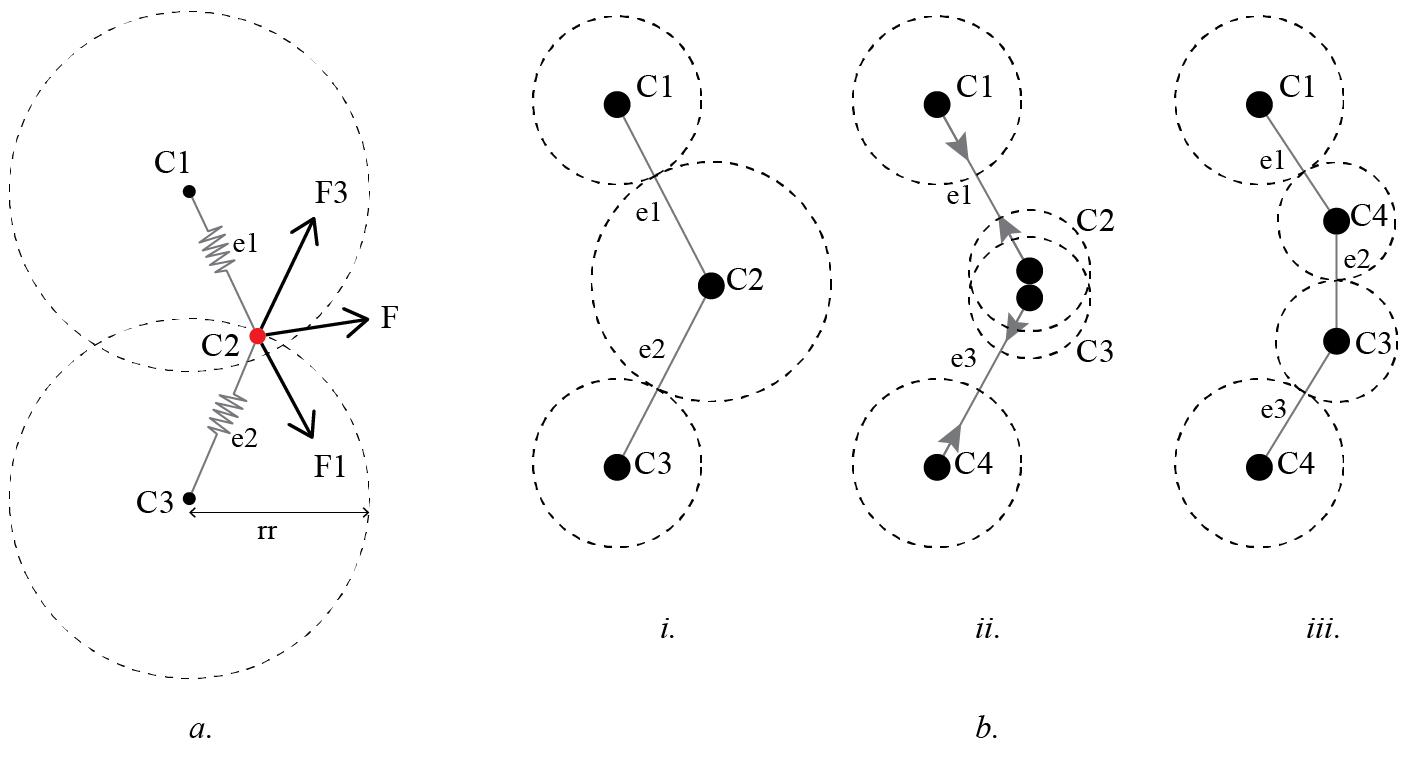}
    \caption{Behaviours of cells ($C$) and edges ($e$). a) Forces acting on $C2$. $F1$ is the sum of forces exerted on $C1$ and $e1$, $F3$ is the sum of the forces exerted by $C3$ and $e2$, and $F$ is the sum of repulsion forces exerted by all the other cells in the system within repulsion radius ($rr$) distance to $C2$. b) Cellular division: (i) $C2$ has reached $\varepsilon_{max}$ (see Sec. \ref{ss:genetic-params}). (ii) $C2$ splits into new cells, $C2$ and $C3$. (iii) $C2$ and $C3$ reach equilibrium. Dotted circles show each cell's energy}
    \label{fig:forces}
\end{figure}

\begin{figure}[hbt]
    \centering
    \includegraphics[width=\textwidth]{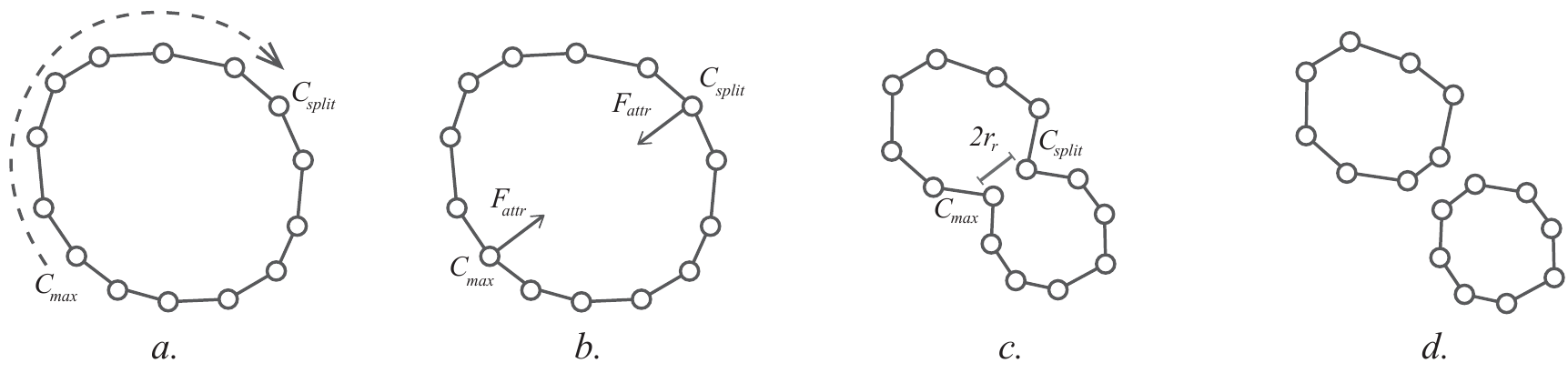}
    \caption{Stages of organism splitting. a) The cell with highest amount of energy $C_{max}$ and $C_{split}$ are defined. b) Attraction forces between $C_{max}$ and $C_{split}$ are applied. c) Distance between $C_{max}$ and $C_{split}$ reaches splitting threshold ($2r_r$)
d) Organism splits, producing two separate and independent organisms.}
    \label{fig:org-division}
\end{figure}

The dynamics of a colony are determined by how its components interact with each other and with the environment. The behaviours of cells and edges - acquiring nutrients, attracting/repelling neighbouring cells (\ref{fig:forces}a), and reproducing (Figure \ref{fig:forces}b) -- determine the shape of organisms as they develop. As a consequence organisms expand, contract, move, change shape and split into multiple organisms, forming a larger colony (Figure \ref{fig:org-division}). The characteristics of these behaviours are defined by the genetic composition of the colony, detailed in Section \ref{ss:genetic-params}.

\subsection{Genetic Parameters}
\label{ss:genetic-params}

The genetic information (\textit{genome}) of a colony consists of the following five alleles:
\begin{itemize}
    \item \textbf{Metabolic rate} ($\eta$), determines how efficient the cells in a colony are at transforming nutrients into energy. A more efficient metabolism comes at the cost of a higher energy consumption.
    \item \textbf{Cell drag coefficient} ($\nu$) represents the surface drag per unit area of cells. Cells with low drag move quickly, making stable configurations more unlikely.
    \item \textbf{Energy Capacity} ($\varepsilon_{max}$) represents the maximum amount of energy a cell can store. As capacity increases, so does the mass of the cells, along with the amount of energy required for movement and metabolism.
    \item \textbf{Edge spring coefficient} ($k$) determines the stiffness of the edges connecting adjacent cells. High $k$ results in rigid organisms. On the contrary, low k results in organisms that move with fluidity, which leads to stretched edges making the transmission of energy between cells less efficient.
    \item \textbf{Energy ratio} ($\rho$) determines the relative size of new organisms after one of them splits into two.
\end{itemize}

\section{Form-finding}
\label{s:form-finding}

To purposefully explore the diverse landscape of design alternatives that our generative system is capable of producing, we developed a set of quantitative representations of design objectives, which are used as fitness functions to drive two evolution-based strategies.

\subsection{Fitness functions}
\label{ss:fitness-functions}

We developed three proof of concept fitness functions, each of which attempts to address a different type of design objective: technical, spatial and aesthetic.

\subsubsection{Printability}
\label{sss:printability}

We base our printability measure on the principles of FDM. The goal is to generate objects that can be printed using a single strand of plastic filament with no additional supports, thus minimising the use of material and eliminating the removal of supporting structures. For this to be feasible two conditions must be met: (i) the smallest diameter of the convex hull of all organisms cannot be smaller than $5mm$ (determined empirically for our specific 3D printer) and (ii) each layer of material has to be supported by the layer below it.

At every timestep during the generation process, the system calculates the minimum diameter of the convex hull of all organisms in a colony, as well as the ratio $P = P_s/P_t$ where $P_s$ is the perimeter of an organism that is in contact with the layer of filament immediately below, and $P_t$ is the total perimeter of the organism. 

\subsubsection{Relative Coverage}
\label{sss:rel-coverage}
This metric determines the area of the largest canopy that can be supported by a given 3D object as a ratio of the  area of the environment. We use a 2D iso-surface approach, as it provides a reasonable approximation. The size of the canopy is calculated by subdividing the environment into equal area square tiles, which are then assigned a support score $m = \sum_{i = 1}^{n} \frac{r_i}{d_i}$, where $n$ is the total number of organisms in the top layer of a 3D shape, $d_i$ is the distance between the centroid of the tile $t_c$ and the centroid of the $i^{th}$ organism and $r_i$ is the distance between the centroid of the $i^{th}$ organism and the point where the line connecting it to $t_c$ intersects the organism’s perimeter. If $m \geq 0.85$ for a tile, it is considered supported. It is important to emphasise that this is an approximation, and that further structural analysis would be required for the development an accurate metric to calculate the load-bearing capabilities of generated objects.

Relative coverage is then calculated as $R_c = R - A$ where $A$ where A is the ratio of the area of the environment covered by the organisms in the first layer, and $R$ is the ratio of supported tiles ($m \geq 0.85$) over the total number of tiles.

\subsubsection{Formal Complexity}
\label{sss:complexity}
This metric attempts to determine if a generated object has physical features that we considered desirable: surface diversity (smooth/rugged), and ``branching''.

To measure surface diversity we consider the convexity of every organism in a colony at every timestep, as well as the quartile coefficient of dispersion of the angles between consecutive edges.

Convexity is calculated as $c = p_h/p$, where $p_h$ is the perimeter of the convex hull of an organism and $p$ is its total perimeter. A highly convex object, such as a circle, should be smooth.

Quartile coefficient is calculated as $Q = (q_3 - q_1)/(q_3 + q_1)$ where $q_1$ and $q_3$ are the first and third quartiles of the array of all angles between consecutive edges $(e_n, e_{n+1})$ sorted by size.

``Branching'' measures the number of times organism splits occur as a colony develops (see Figure \ref{fig:org-division} for reference). We then calculate a splitting score $S = d^{(1/(1+n_s))}$ where $d$ is a constant and $n_s$ is the total number of splits.

The overall complexity of a generated object is calculated as the equal-weighted sum of the three metrics described above: $C = (c + Q + S)/3$.

\subsection{Exploration strategies}
\label{ss:exploration-strategies}

The general aim of this research was to develop a system that enables designers to explore diverse landscapes of designs using a combination of domain knowledge and intuition, in search of functional alternatives with novel/unexpected features.

We approximate the desired characteristics with our fitness functions (see Section \ref{ss:fitness-functions}). However, due to the uncertain nature of some design goals -- especially those related to perceptual attributes, like aesthetic complexity -- we regard optimisation processes as instrumental for exploration, rather than tools for the efficiently searching for the best alternative. This is why we used a hybrid, two-stage exploration strategy that combines a powerful optimisation algorithm with a method that introduces some intuitive aspects of creative exploration.

\subsubsection{Evolution Strategy}

Firstly we leverage the optimisation capabilities of the Covariance Matrix Adaptation Evolution Strategy (CMA-ES) \citep{hansen2001completely}, a state of the art population-based optimisation algorithm that works by randomly sampling alternatives in the area of the fitness landscape close to other successful alternatives, using a distribution that is updated every generation based on the previous generation’s most successful individuals. Despite the fact that the efficiency of evolutionary algorithms for design optimisation has been contested \citep{wortmann2017genetic}, we selected CMA-ES for two specific reasons: (i) its adaptive capabilities make it suitable for the exploration of multimodal fitness landscapes, a characteristic that our generative system displayed during preliminary testing, and (ii) it works well with low-dimensional floating point vectors as genomes, as it does not rely on crossover to generate new genotypes.

An argument could be made against the suitability of CMA-ES to address a multi-objective problem. We believe that, given the imprecise nature of some of the design objectives being pursued, it is beneficial to address them individually, and then allow the user to fill-in the shortcomings of the formal implementation of said objectives via a procedure that enables intuitive selection.

\subsubsection{Linear Interpolation Search}

For the second stage of our method we produce a series of new genotypes by interpolating between the genotypes of two user-selected CMA-ES evolved objects. To do this we calculate the Euclidean distance between them, and then divide it by the number of desired ``in-between'' objects. These genotypes generate new objects that can be examined by the user, both visually and by evaluating their properties using the measures described in Section \ref{ss:fitness-functions}. This allows the user to optimise to taste, while preserving measurable fitness, fine-tuning the details that formal fitness measures are unable to capture sufficiently.

We believe that this method, coupled with the more traditional use of CMA-ES, has the potential to expand the creative capabilities of designers, as it enables the user of the system to introduce their own aesthetic preference as part of a process that is also capable of maintaining performance standards.

\section{Experiments}
\label{s:experiments-results}

\subsection{Experimental setup}

The first stage of our experiments involved the progressive transformation of the genome vector ($\eta, \nu, \varepsilon_{max}, k, \rho$) using CMA-ES. We did $5$ evolutionary runs of $150$ generations each, using a population of $\lambda = 40$ individuals at each generation. From each population, a set of $2$ \textit{parents} was selected to define the distribution for the sampling of the new population. For the generation of individual objects we set the environment size to $(600 \times 600)$ units. Each object was initialised as a colony of $4$ organisms, located on the midpoints of the diagonal lines connecting the centre of the environment and its vertices. Colonies were developed for 200 timesteps (with 100 timesteps of warmup), resulting in objects 200 layers high. For each evolutionary run the environment was initialised with a different random seed, in order to obtain variation in form via the random distribution of nutrient sources. For the second stage of the experiments two individuals generated on identical environments were selected based on their aesthetic characteristics and fitness. These individuals were used to run a linear interpolation between genome vectors, generating $99$ in-between new individuals which were then evaluated for fitness, as well as by visual inspection.

\subsection{Results}

\begin{figure}[hbt]
    \centering
    \includegraphics[width=\textwidth]{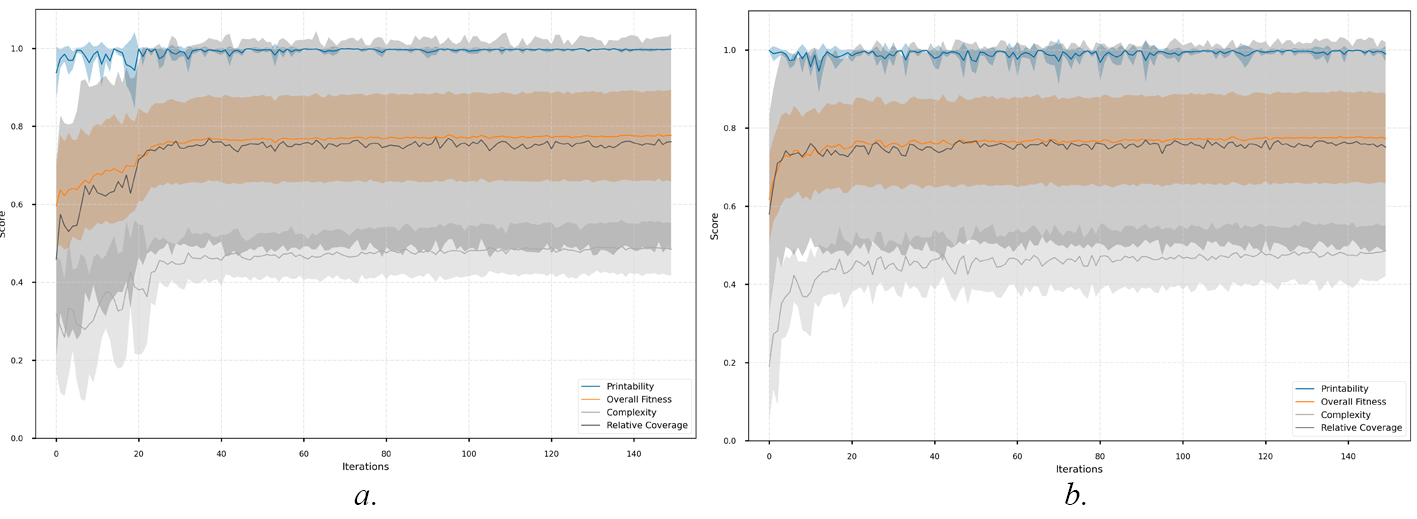}
    \caption{Plots of evolutionary runs showing mean scores for the best individual at every generation averaged over 5 runs. Shaded areas illustrate the variance. In a) we optimise for full fitness, and in b) we optimise for relative coverage.}
    \label{fig:plot_overall}
\end{figure}

The first stage of experiments tested the ability of the CMA-ES algorithm to find designs with specific characteristics. Figure \ref{fig:plot_overall} shows the results when optimising for overall fitness (averaging printability, complexity and relative coverage) (a), and when optimising for relative coverage (b). Even though both graphs show similar results, it is possible to make interesting observations. For instance, in (a) printability (blue) quickly reaches its optimum value and exhibits little variance, whereas complexity (light grey) and relative coverage (dark grey) exhibit a fairly dramatic relative increase and follow remarkably similar trajectories, but do not reach high values. This suggests that once the system finds a highly printable alternative, it becomes difficult for it to find objects that perform well in other metrics. However, an assessment of graph (b) indicates that the system struggles to reach high scores for relative coverage in general.

\begin{figure}[htb]
    \centering
    \includegraphics[width=\textwidth]{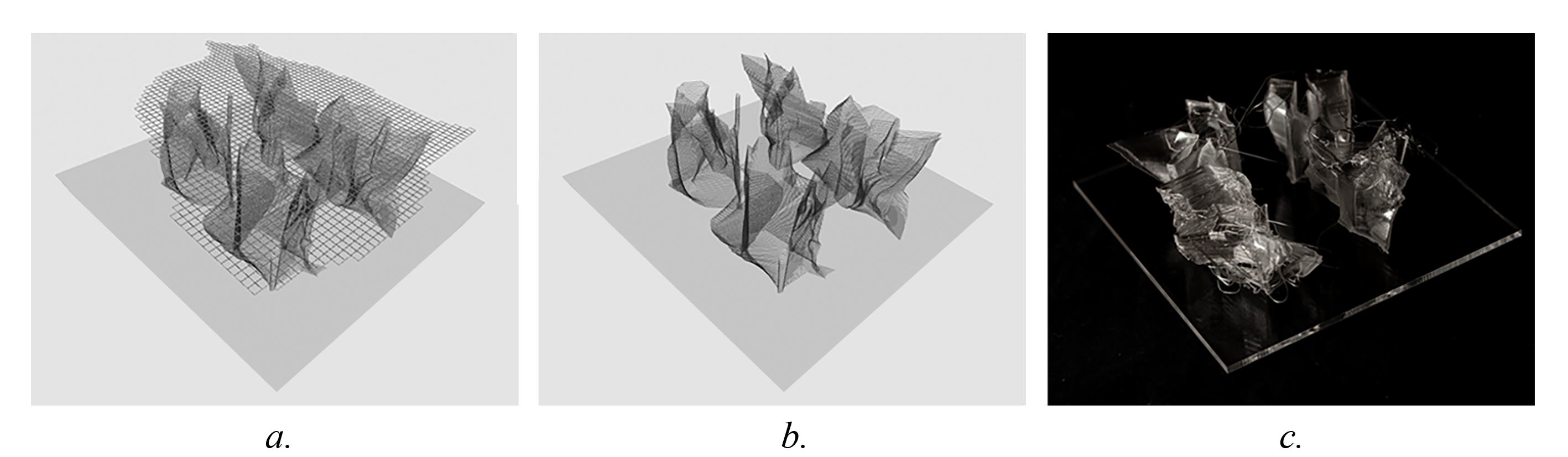}
    \caption{Object obtained by optimising coverage. a) Contour 3D rendering (with projected canopy), b) Contour 3D rendering (no canopy) and c) 3D printed version.}
    \label{fig:results_coverage}
\end{figure}

To further understand the characteristics of the objects that the system can generate, we digitally render the highest performing individual for each generation. We then manually select the ones that exhibit interesting characteristics or are highly fit. Figure \ref{fig:results_coverage} shows an example obtained from one of the runs in which we optimised for relative coverage. The pavilion-like shape of the object exhibits a set of complex structures capable of supporting a canopy that covers about $60\%$ of the environment. However, a printability $P < 1.0$ renders the object unprintable.

\begin{figure}[htb]
    \centering
    \includegraphics[width=\textwidth]{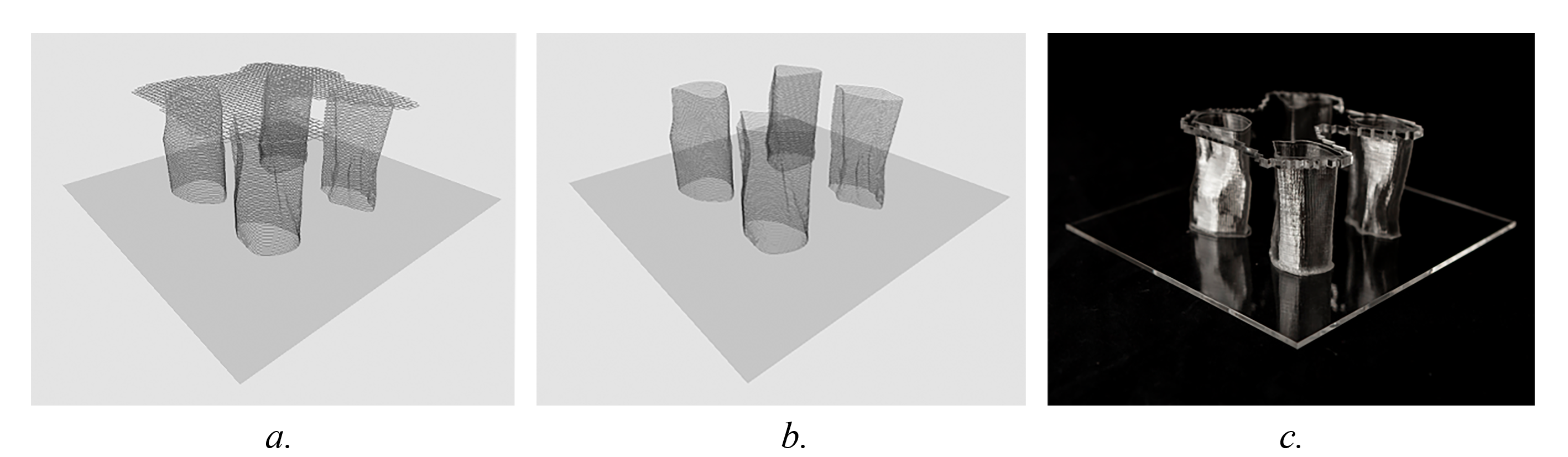}
    \caption{Object obtained by optimising for full fitness. a) Contour 3D rendering (with projected canopy), b) Contour 3D rendering (no canopy) and c) 3D printed version.}
    \label{fig:results_full_fitness}
\end{figure}

The object shown in Figure \ref{fig:results_full_fitness}, in contrast, exhibits simple geometry, low relative coverage and high printability, allowing for a successful 3D print (to which we added a laser-cut canopy for reference). The lack of formal complexity is not necessarily negative, as from an aesthetic point of view the object appears structurally robust and its components are easily discernible. This is why we selected it as the second end point for the interpolation procedure.

\begin{figure}[hbt]
    \centering
    \includegraphics[width=\textwidth]{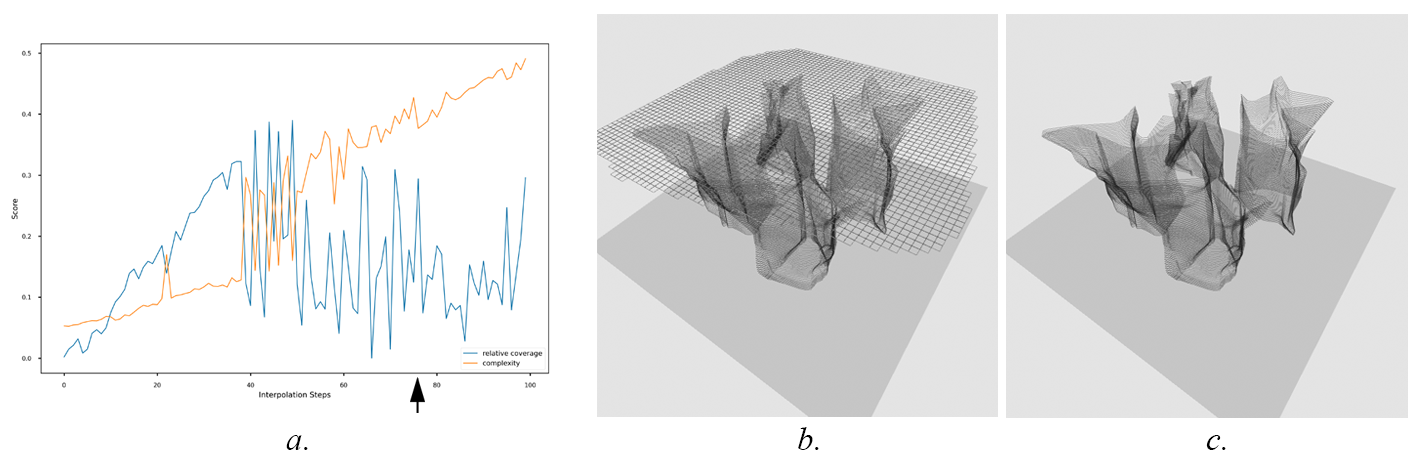}
    \caption{Interpolation results. a) Relative coverage (blue) and complexity (orange) scores through interpolation. b) Selected 3D object (with projected canopy). c) Selected 3D object.}
    \label{fig:results_interpolation}
\end{figure}

Figure \ref{fig:results_interpolation} shows the outcomes of the exploration stage. In the graph (a) it is possible to observe unexpected behaviours of the fitness values for complexity and relative coverage as we interpolate between objects. The overall trend shows complexity increasing as coverage decreases. However the random value spikes and drops between steps 40 and 50 suggest that the fitness landscape is not smooth, making it harder for the optimisation algorithm to find the overall best solutions. Nevertheless, this can be fertile ground for creative opportunities, as the user of our system may encounter unexpected attributes in objects found in those areas of the landscape. For instance, the object depicted in (b) and (c) -- iteration 78 -- exhibits similar characteristics to the object in Figure \ref{fig:results_coverage}, yet, on closer inspection, it is also possible to see that its contours are smoother, which indicates that some features of the object in Figure \ref{fig:results_full_fitness} have been captured to produce an object \textit{in-between}.

\section{Discussion and conclusion}
\label{s:discussion-conclusion}

This paper introduced a novel generate/explore digital design system inspired by natural and digital-fabrication processes. The method enables users to produce and purposefully search through a vast landscape of design alternatives. Our goal was to examine the capability of our novel stochastic agent-based generative system to produce a diverse landscape of design alternatives, as well as how the integration of automated and ``manual'' selection mechanisms can widen the range of creative opportunities for designers.

Via the implementation of a two-stage approach that considers optimisation and guided exploration through interpolation – searching in-between designs – we took initial steps toward the development of a system that enables users to uncover areas of the design landscape that hold unexpected design solutions. Our results indicate that these systems have the potential to expand the creative capabilities of the designers who use them.

The strength of our system lies on its capability to combine the optimisation power of the CMA-ES algorithm with human creative judgment. This opens up areas of the design landscape that are extremely difficult to discover by either approach if used separately. We foresee many opportunities to continue this work, beyond the further development of the system’s components. This includes the exploration of new methods for generation and optimisation, and the implementation of a user interface. Another avenue for investigation is the development of fitness functions informed by human perception, to evaluate hard-to-measure features, like ``spatial continuity'' or ``openness''. This could be achieved by, e.g., generating a large database of categorised objects to train a classifier model for the recognition of features in newly generated objects.

\printbibliography
\end{document}